\documentclass{article}
\usepackage[T1]{fontenc}
\usepackage[utf8]{inputenc}
\usepackage{ismir} % Remove the "submission" option for camera-ready version
\usepackage{amsmath,cite,url}
\usepackage{graphicx}
\usepackage{color}

\usepackage{multirow}
\usepackage{booktabs}
\usepackage{cellspace} 

\usepackage{enumitem}
\setlist[itemize]{itemsep=2pt, parsep=2pt, topsep=2.5pt, leftmargin=1.5em}
\setlist[enumerate]{itemsep=2pt, parsep=2pt, topsep=2.5pt, leftmargin=1.5em}

\makeatletter
\renewcommand{\paragraph}[1]{%
  \vspace{0.25\baselineskip}%
  \noindent\textbf{\textit{#1}}\ %
}
\makeatother

\title{A Dataset and Benchmark for Optical Music Recognition of String Quartet Scores}

\multauthor
  {Dongmin Kim$^1$ \hspace{1cm} Brian Liu$^2$ \hspace{1cm} Jose J. Valero-Mas$^3$ \hspace{1cm} Dasaem Jeong$^1$}
  {    
      \\
      $^1$ Music \& Arts Learning (MALer) Lab, Sogang University, Seoul, South Korea\\
      $^2$ Independent Researcher\\
      $^3$ Pattern Recognition and Artificial Intelligence Group, University of Alicante, Alicante, Spain\\
      {\tt\small dasaemj@sogang.ac.kr}
  }

\def\authorname{D. Kim, B. Liu, J. J. Valero-Mas, and D. Jeong}
\begin{document}

\maketitle

%-----------------------------------------------------
\begin{abstract}
Optical music recognition (OMR) transcribes music scores into digital formats.
While the field has advanced significantly on monophonic and piano-form scores, multi-part score transcription remains underexplored, largely due to the absence of a suitable dataset.
We introduce {OpenScore String Quartet} for {Optical Music Recognition} (OSSQ-OMR), the first dataset dedicated to multi-part OMR. Built on the OpenScore String Quartet corpus, OSSQ-OMR pairs digitally encoded scores with their original scanned editions from IMSLP, with all images visually aligned to their transcriptions.
The dataset is released with score images at system and staff levels, and paired transcriptions in three encoding formats: Extended Linearized MusicXML (LMXE), \texttt{**kern}, and ABC. 
In total, OSSQ-OMR contains 24{,}544 system images and 98{,}172 staff images drawn from 116 string quartet scores.
We accompany the dataset with a benchmark protocol and baseline results from two representative OMR models, evaluated across four random score-level splits with mutually exclusive test sets. 
Baselines reach OMR-NED as low as 3.6\% on synthetic and 5.9\% on scanned inputs; results reveal substantial effects of encoding and segmentation choices, with the LSTM-based baseline degrading on scanned inputs roughly 2.6$\times$ less than the Transformer-based baseline.

\end{abstract}
%-----------------------------------------------------

\begin{figure*}[!ht]
    \centering
    \includegraphics[
        width=.8\textwidth,
        alt={Score excerpts from OSSQ-OMR illustrating the visual diversity of the dataset across publishers, eras, and source conditions. The samples span typeset full scores from major 19th- and 20th-century publishers (Breitkopf \& H\"artel, Ernst Eulenburg, Trautwein, and others), 18th-century engraving conventions, modern digital re-engravings, and handwritten manuscripts.}
    ]{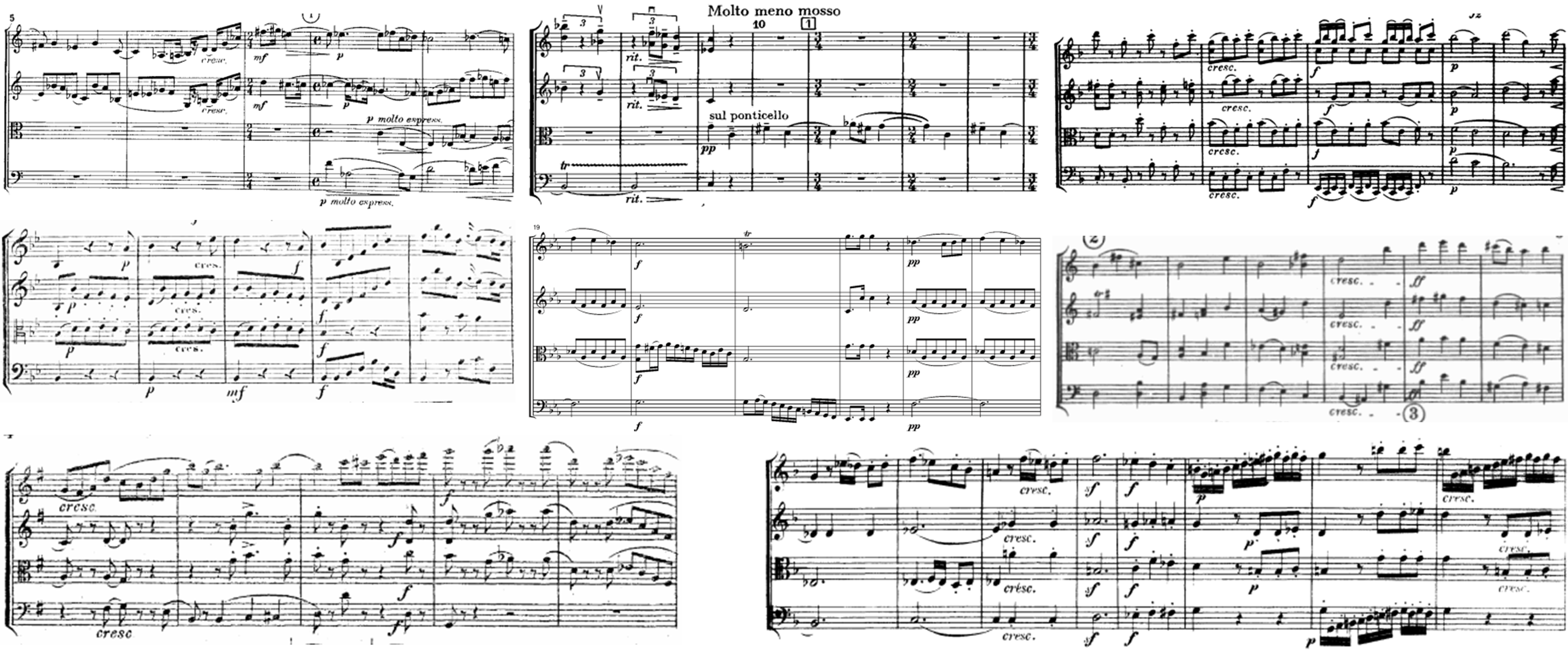}
    \caption{Score excerpts from OSSQ-OMR illustrating the visual diversity of the dataset across publishers, eras, and source conditions. The samples span typeset full scores from major 19th- and 20th-century publishers (Breitkopf \& H\"artel, Ernst Eulenburg, Trautwein, and others), 18th-century engraving conventions, modern digital re-engravings, and handwritten manuscripts.}
    \label{fig:diversity}
\end{figure*}
%-----------------------------------------------------
\section{Introduction}\label{sec:introduction}
%-----------------------------------------------------
Optical music recognition (OMR) refers to the field that studies computational methods for automatically transcribing music score images into structured digital formats~\cite{CalvoZaragoza:ACMCS:2020}. It plays a key role in the preservation of musical cultural heritage while enabling its exploitation and analysis through Music Information Retrieval (MIR) techniques based on symbolic representations~\cite{jones2008optical,miranda2021handbook}.

Building on their success in image-related tasks, OMR research increasingly relies on deep learning frameworks due to their strong representation learning capabilities~\cite{omrsurvey,castellanos2025deep}. In particular, current strategies focus on neural end-to-end (or holistic) approaches that formulate score recognition as a single task, either at the staff level~\cite{CastellanosCalvoZaragozaInesta:ISMIR:2020,CastellanosGarridoMunozRiosVilaCalvoZaragoza:ESWA:2022,HarteltEipertPuppe:AS:2024,CastellanosMartinezEstesoGalanCuencaGallego:ICDAR:2024}, typically preceded by staff-line segmentation, or, more recently, at the full-page level~\cite{rios2023end,rios2024sheet,rios2024sheet2,rios2025implicit,rios2026end}. In both cases, encoder–decoder architectures are commonly adopted, combining convolutional encoders for feature extraction with sequential decoders (e.g., recurrent, transducer, or Transformer-based models) to capture structural dependencies~\cite{baro2017optical,calvo2017end,wen2022sequence,Edirisooriya:ISMIR:2021,rios2026end}.

Most neural-based OMR developments have focused on monophonic music, largely due to the abundance of historical sources (e.g., plainchant) and the straightforward recognition formulation. Representative works and datasets include PrIMUS~\cite{CalvoZaragoza:ISMIR:2018}, Capitan~\cite{calvo2016two}, Seils~\cite{ParadaCabaleiroBatlinerSchuller:ISMIR:2019}, FMT~\cite{RosFabregasMunnichRizo:MEI:2021}, and Guatemala~\cite{ThomaeCummingFujinaga:DfLM:2022}. More recent work has begun to address more complex textures, such as piano-form scores, as exemplified by the GrandStaff collection~\cite{rios2023grandstaff,rios2026end}. However, progress on other textures, such as \emph{multi-part} scores---where multiple voices are performed simultaneously by different instruments---remains limited, largely due to the scarcity of suitable datasets.

To address this gap, we introduce the OpenScore String Quartet for Optical Music Recognition (OSSQ-OMR)\footnote{\raggedright The dataset and benchmark code are publicly available at \url{https://github.com/MALerLab/string-quartet-omr-benchmark}.}, the first dataset dedicated to multi-part OMR, targeting string quartet repertoire. OSSQ-OMR extends the OpenScore String Quartet corpus~\cite{gotham:ossq} along key OMR axes: (i) each digital score is paired with its corresponding IMSLP scan~\cite{imslp}; (ii) score images are provided at system and staff levels; and (iii) digital scores are released in three encoding formats: Extended Linearized MusicXML (LMXE), \texttt{**kern}, and ABC. The resulting dataset comprises 13{,}240 synthetic and 11{,}304 scanned system images, alongside 52{,}960 synthetic and 45{,}212 scanned staff images, derived from 116 string quartet scores.

Compared to piano-form notation, these scores comprise four independent instrumental parts per system, substantially increasing complexity and introducing challenges such as wide pitch ranges, frequent ledger lines, and multiple clef changes across and within movements.

Finally, we also provide a benchmark protocol and baseline results using two representative OMR architectures, evaluated across all segmentation levels and encoding configurations. The resulting error rates---below 4\% and 6\% on synthetic and scanned inputs, respectively---confirm that the task is feasible and show that both encoding and segmentation choices significantly impact performance, highlighting open challenges in multi-part OMR.

%-----------------------------------------------------
\section{The OSSQ-OMR dataset}\label{sec:dataset}
%-----------------------------------------------------

\subsection{Source corpus and scan collection}\label{subsec:corpus}

We build the dataset from the OpenScore String Quartet corpus~\cite{gotham:ossq}, comprising 116 string quartet scores encoded as 122 MuseScore files (with two multi-movement works split by movement), along with rendered image versions. To extend it, we retrieve the corresponding scanned scores from IMSLP~\cite{imslp} and categorize them as follows:
\begin{itemize}
    \item \textbf{Full score} (93 scores, 80\%): all four parts are shown on a single system. These are predominantly typeset editions, except for one digitally engraved score (Arriaga's String Quartet No.\,3 in E-flat major), and constitute the primary OMR targets.

    \item \textbf{Part-book} (17 scores, 15\%): single-instrument parts, reflecting the 18th- and 19th-century practice of publishing chamber music as separate booklets.

    \item \textbf{Manuscript} (6 scores, 5\%): handwritten manuscripts with varying quality, all by women composers. %
\end{itemize}

OSSQ-OMR inherits the compositional diversity of its source corpus, comprising 116 scores from 47 composers spanning publication dates from 1773 to 1994 (plus one later re-engraving). The dataset covers 32 publishers and six autograph manuscripts, with the three largest publishers (Breitkopf \& H\"artel, Ernst Eulenburg, and Trautwein) accounting for 56\% of the collection. Figure~\ref{fig:diversity} shows representative excerpts illustrating the visual range of engraving styles, layouts, and source conditions in the dataset.

\subsection{Visual alignment and corrections} \label{subsec:alignment}
\begin{figure*}[!ht]
    \centering
    \includegraphics[
        width=.8\textwidth,
        alt={Examples of visual mismatches between OpenScore digital encodings and their corresponding scanned editions, motivating the visual-alignment process described in Section "Visual alignment and corrections". In each subfigure, the top system shows the scanned source and the bottom system shows the OpenScore rendering before correction; differences are marked in red. (a) Mozart, String Quartet No.\,17 in B-flat major, K.\,458: the scanned edition omits a measure present in OpenScore (likely a printing error); to align visually, we remove the measure from the encoding. (b) Glinka, String Quartet: the scanned source switches from tenor to bass clef in the third measure, while the OpenScore encoding uses a bass clef throughout.}
    ]{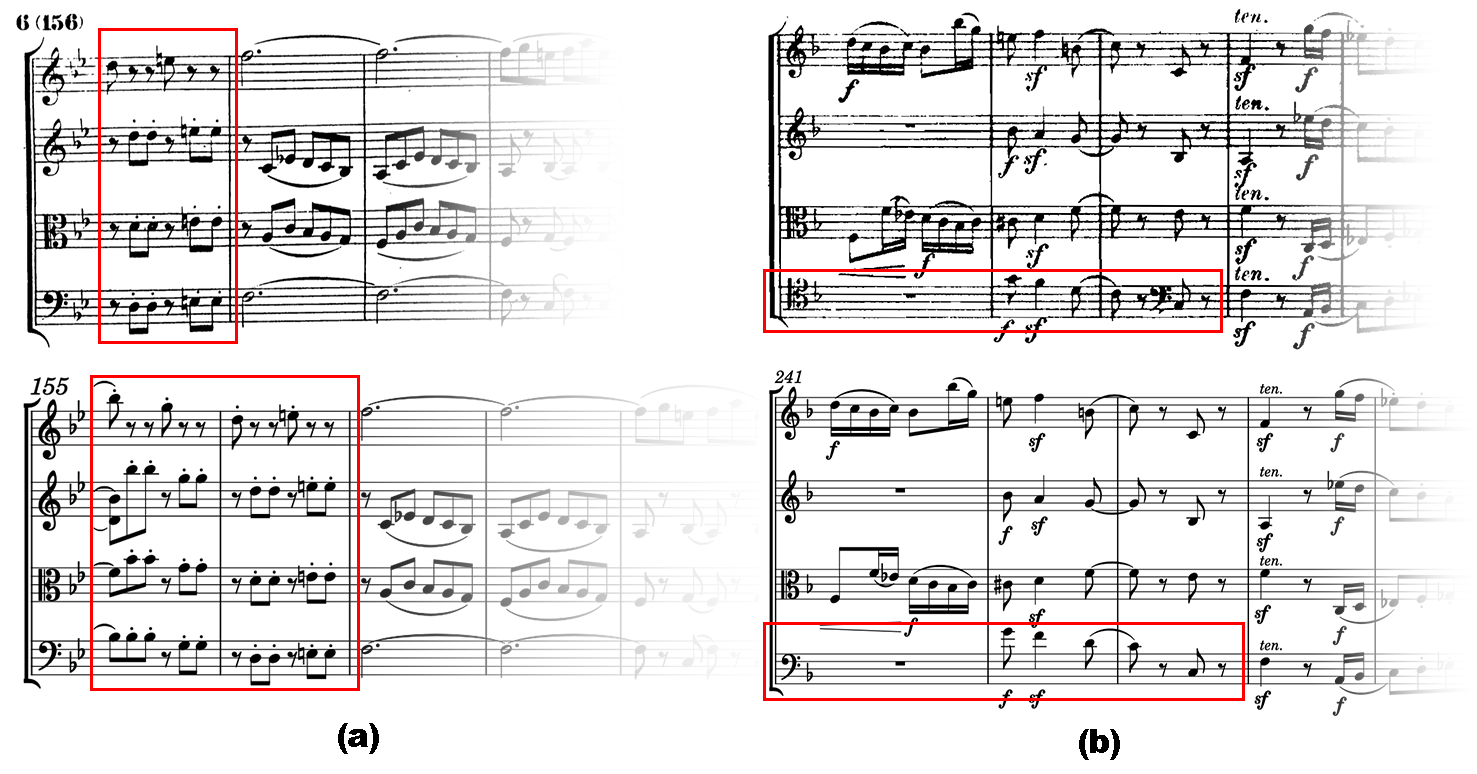}
    \caption{Examples of visual mismatches between OpenScore digital encodings and their corresponding scanned editions, motivating the visual-alignment process described in Section~\ref{subsec:alignment}. In each subfigure, the top system shows the scanned source and the bottom system shows the OpenScore rendering before correction; differences are marked in red. \textbf{(a)} Mozart, String Quartet No.\,17 in B-flat major, K.\,458: the scanned edition omits a measure present in OpenScore (likely a printing error); to align visually, we remove the measure from the encoding. \textbf{(b)} Glinka, String Quartet: the scanned source switches from tenor to bass clef in the third measure, while the OpenScore encoding uses a bass clef throughout.}
    \label{fig:corrections}
\end{figure*}

OpenScore's editorial process normalizes notation toward clean, performer-ready editions, smoothing inconsistencies in the source scans, including original print errors and outdated typesetting conventions~\cite{gotham:ossq}. For OMR, however, we require digital encodings whose rendered images match the corresponding scans at the token level. Figure~\ref{fig:corrections} shows two representative mismatches between OpenScore encodings and their scanned sources. To address such gaps, we re-edit each in-scope score so that the MuseScore rendering aligns visually with its IMSLP scan.

\paragraph{Process.} The alignment was performed by the authors of the work, who iteratively edited, reviewed, and validated the results. This process was exclusively applied to the 93 full-score scans and spanned over 100 hours.

\paragraph{Auditability.} We release the corrected MuseScore sources as a Git repository whose history records each edit as a commit on top of the original OpenScore encoding, making every layout and notation change individually auditable.

\paragraph{Scope of edits.} Edits are grouped into two categories. \textit{Layout edits} adjust system, page, and section breaks to match the scanned layout. \textit{Notation edits} modify musical content, including clefs, stem directions, slurs and ties, tremolos, beaming, and articulations. All changes are applied directly to the MuseScore files, ensuring that a single source produces both the aligned rendering and the released symbolic transcription.

\paragraph{Extent of corrections.} Of the 93 full-score scans, 78 required both layout and notation edits, 6 required only layout edits, 8 required only notation edits, and one required no edits at all. Table~\ref{table:corrections} summarizes the most frequently edited symbol categories. 

\begin{table}[!ht]
\centering
\begin{tabular}{lrr}
\toprule
\textbf{Category} & \textbf{\# files} & \textbf{Edits} \\
\midrule
System breaks       & 81 & 984 \\
Page breaks         & 81 & 939 \\
Beaming             & 69 & 1{,}262 \\
Slurs               & 37 & 1{,}948 \\
Clef markings       & 34 & 786  \\
Articulations       & 16 & 1{,}080 \\
Notes (pitch edits) & 21 & 7{,}183 \\
\bottomrule
\end{tabular}
\caption{Detail of the most common notation edits performed on the full-score pieces.}
\label{table:corrections}
\end{table}

\subsection{Image segmentation}\label{subsec:segmentation}

OSSQ-OMR provides score images at two levels of granularity: system and staff. System- and staff-level images are produced by a three-stage pipeline of YOLOv8~\cite{yolov8} models, shared across synthetic and scanned inputs.

\paragraph{Image sources.} For synthetic images, we render each MuseScore file to PDF using MuseScore 3.6.2~\cite{musescore} with A4 page size and the default engraving style, then rasterize each PDF page to PNG at 300 DPI. For scanned images, we use the IMSLP PDFs directly, rasterized to PNG at their native resolution. %

\paragraph{Segmentation pipeline.} Three YOLOv8 models~\cite{yolov8} are sequentially applied:
\begin{enumerate}
\item \textbf{System bounding-box detection.} The first model detects systems on a page-level image. We crop each detection to produce a system-level image.
\item \textbf{Staff-height regression.} The second model regresses the staff height in pixels on a system-level image. We resize each system image uniformly so that noteheads (one quarter of staff height) are approximately 4--5 pixels in height, matching the input scale expected by downstream OMR models.
\item \textbf{Staff bounding-box detection.} The third model detects individual staves on a resized system-level image.
\end{enumerate}

\paragraph{Training data.} 
We use the staff-height regression model from~\cite{umust2026} as-is, train the system-detection model from scratch on a combined set of 1{,}804 pages from the reference work and 47 OSSQ-OMR pages, and train a new staff-detection model on 1{,}440 OSSQ-OMR system images. Annotations were reviewed by the authors.

\subsection{Symbolic encoding formats} \label{subsec:encoding}

Each OSSQ-OMR score is released with symbolic transcriptions as plain text in three encoding formats:

\paragraph{LMXE.} Extended Linearized MusicXML (LMXE) extends Linearized MusicXML~\cite{mayerZeus} to support multi-part scores while preserving its compact, sequence-friendly representation. It introduces (i) explicit \texttt{part} tokens and a score-type token (\texttt{single}, \texttt{multi}, \texttt{grandstaff}) to unify different score types under a single vocabulary, and (ii) a measure-wise encoding order that nests parts within measures, improving robustness to decoding errors by ensuring partial outputs remain reconstructable.

\paragraph{\texttt{**kern} and ABC.} The Humdrum \texttt{**kern} format~\cite{kern,rios2023end,rios2024sheet} organizes voices and staves in tab-separated spines, a comparatively rigid structural constraint on valid encodings; ABC notation~\cite{walshaw2021abc,legato2026} instead uses ASCII mnemonics with little structural scaffolding, and its textual simplicity suits general-purpose language models.

\paragraph{Conversion pipeline.} All encodings are derived from a shared pipeline starting from each MuseScore file:
\begin{enumerate}
\item MuseScore files are exported to MusicXML using MuseScore~\cite{musescore}. The resulting MusicXML undergoes a cleansing stage to remove invisible elements and their associated backup/forward directives, which would otherwise produce visually absent but symbolically present content.
\item The cleaned MusicXML is converted to system-wise LMXE using a custom tool based on~\cite{mayerZeus}.
\item From system-wise LMXE, we derive system-wise \texttt{**kern} and system-wise ABC by converting back to MusicXML and then forward using \texttt{converter21}.\footnote{\url{https://github.com/gregchapman-dev/converter21}}
\item Part-wise encodings are derived as follows: part-wise LMXE is produced directly from system-wise LMXE using the custom tool; part-wise \texttt{**kern} is produced from system-wise \texttt{**kern} using \texttt{kernpy}; part-wise ABC is produced by converting part-wise LMXE to MusicXML and then to ABC via \texttt{converter21}.
\end{enumerate}

\paragraph{Verification and exclusions.} Each conversion step except the initial MuseScore export and the \texttt{**kern} part-wise split is verified by round-trip conversion back to MusicXML, with the round-trip output compared to its input using OMR-NED~\cite{omr-ned}. Samples for which any of the structural or major musical error categories \{measure insert/delete, staff insert/delete, staff group, barline, key signature, clef, tie, note, dot\} are detected on round-trip are excluded from the release, as are samples for which any conversion step raises an error. Approximately 5\% and 21\% of system-wise samples are lost during \texttt{**kern} and ABC conversion, respectively. Part-wise data inherits the same exclusion mask. Exclusions are applied at the sample (system or part) level, not the score level: every in-release score contributes at least some systems and parts, though individual systems may be absent from one or more encoding tracks.

\subsection{Dataset statistics} \label{subsec:stats}
After segmentation and encoding conversion, OSSQ-OMR releases 24{,}544 system images and 98{,}172 staff images drawn from 116 synthetic and 93 scanned string quartet scores, paired with symbolic transcriptions in three encoding formats. Table~\ref{tab:release-counts} reports the final per-track release counts. Exclusions are per-track: a sample absent from one encoding may still be present in others.

\begin{table}[!ht]
\centering
\small
\begin{tabular}{ll|rrrr}
\toprule
\textbf{Source} & \textbf{Level} & \textbf{Raw} & \textbf{LMXE} & \textbf{Kern} & \textbf{ABC} \\
\midrule
\multirow{2}{*}{Synthetic} & System & 13{,}241 & 13{,}240 & 12{,}505 & 10{,}434 \\
                           & Part   & 52{,}960 & 52{,}960 & 50{,}020 & 41{,}728 \\
\midrule
\multirow{2}{*}{Scanned}   & System & 11{,}305 & 11{,}304 & 10{,}691 & \phantom{0}8{,}850 \\
                           & Part   & 45{,}212 & 45{,}212 & 42{,}760 & 35{,}388 \\
\bottomrule
\end{tabular}
\caption{OSSQ-OMR release counts. ``Raw'' is the post-segmentation, pre-conversion sample count. Track columns report counts after per-track conversion verification removes samples with structural or major musical round-trip errors (Section~\ref{subsec:encoding}). LMXE is derived without conversion to a third format and therefore has near-full coverage.}
\label{tab:release-counts}
\end{table}

%-----------------------------------------------------
\section{Benchmark protocol}\label{sec:method}
%-----------------------------------------------------
We accompany OSSQ-OMR with benchmark experiments to analyze the capabilities of current state-of-the-art OMR models on multi-part scores. The remainder of the section describes the experimental details: baseline models, tokenization schemes, data splits, and evaluation metrics.

\subsection{Baseline models}
We evaluate two baseline architectures representative of the state of the art in piano-form OMR:

\textbf{Zeus}~\cite{mayerZeus}. An LSTM-based sequence-to-sequence model with a ResNet-like~\cite{resnet} convolutional encoder, designed for linearized MusicXML transcription. In our experiments, we increase the encoder and decoder LSTM hidden sizes to 256.

\textbf{Sheet Music Transformer} (SMT)~\cite{rios2024sheet}. An encoder–decoder model combining a ConvNeXt~\cite{convnext} vision encoder with a Transformer decoder, originally developed for \texttt{**kern}-based OMR. No architectural modifications are applied.

\subsection{Tokenization schemes}
\label{subsec:tokenizations}
We evaluate nine tokenization schemes spanning the three release encoding formats. For each scheme, we report the maximum sequence length cap (system-wise / part-wise tokens) used during data filtering (Section~\ref{subsec:splits}):

\begin{itemize}
\item \textbf{LMXE} (800 / 430): space-separated tokens from the LMXE converter (Section~\ref{subsec:encoding}).
\item \textbf{LMXE-P} (790 / ---): an ablation variant with part-major encoding order (analogous to MusicXML's \texttt{partwise} layout); used only at the system level to justify the design choice in Section~\ref{subsec:encoding}.
\item \textbf{EKERN} (1{,}100 / 390)~\cite{rios2024sheet2}: a \texttt{**kern} tokenization splitting symbols by graphical meaning, with a delimiter and canonical ordering. Produced by \texttt{kernpy}~\cite{kernpy}.
\item \textbf{BEKERN} (1{,}400 / 540)~\cite{rios2023end, rios2024sheet}: a finer-grained variant of EKERN that decomposes each note into its minimum semantic elements; smaller vocabulary, longer sequences. Also produced by \texttt{kernpy}.
\item \textbf{CABC} (1{,}170 / 500): character-level tokenization of ABC.
\item \textbf{ABC-BPE} (340 / 160): byte-pair encoding~\cite{bpe} of ABC with a 4{,}096-token vocabulary, matching~\cite{legato2026}.
\item \textbf{ABC-BPE-1024} (430 / 210), \textbf{ABC-BPE-512} (500 / 260), \textbf{ABC-BPE-256} (640 / 340): ablation variants with smaller BPE vocabularies.
\end{itemize}

\subsection{Data filtering and splits}
\label{subsec:splits}

Two processes are initially applied to the data to filter out unsuitable elements:
\begin{itemize}
    \item Samples whose system-level image exceeds 256 pixels in height (after the resizing described in Section~\ref{subsec:segmentation}) are excluded, as they exceed the input constraint of the baseline models. 
    \item Samples whose encoded sequence exceeds the per-encoding length cap listed in Section~\ref{subsec:tokenizations} are excluded; a sample is dropped globally if it fails the cap for any of the nine encodings, so that every benchmark configuration uses the same pool of samples. Per-encoding caps were chosen to balance the available compute budget against sample coverage: smaller caps reduce training and decoding time but exclude more samples, particularly for verbose encodings. The resulting pool is the intersection of valid samples across all encodings. 
\end{itemize}

We split the remaining samples at the score level to prevent cross-split leakage: all systems and parts belonging to a given score are kept in a single split. Four random splits were generated, with mutually exclusive test sets across splits. Table~\ref{tab:splits} reports per-split sample-count statistics; test sets are additionally divided between synthetic and scanned sources. All reported baseline numbers are averaged across the four splits.

\begin{table}[!ht]
\centering
\small
\begin{tabular}{lrr}
\toprule
\textbf{Partition} & \textbf{System} & \textbf{Part} \\
\midrule
Train        & 13{,}552\,$\pm$\,128 & 54{,}207\,$\pm$\,513 \\
Valid        & \phantom{0}1{,}611\,$\pm$\,\phantom{00}1   & \phantom{0}6{,}441\,$\pm$\,\phantom{00}4 \\
Test (synth) & \phantom{00}\phantom{0}875\,$\pm$\,130 & \phantom{0}3{,}499\,$\pm$\,518 \\
Test (scan)  & \phantom{00}\phantom{0}957\,$\pm$\,\phantom{00}8   & \phantom{0}3{,}825\,$\pm$\,\phantom{0}31 \\
\midrule
Total (synth)  & 10{,}016\phantom{\,$\pm$\,000} & 40{,}064\phantom{\,$\pm$\,000}       \\
Total (scan)   & 8{,}509\phantom{\,$\pm$\,000}  & 34{,}032\phantom{\,$\pm$\,000}        \\
\bottomrule
\end{tabular}
\caption{Per-split sample-count statistics (mean\,$\pm$\,std across the four splits) at system and part levels. Test sets across splits are mutually exclusive at the score level.}
\label{tab:splits}
\end{table}

\begin{table*}[!ht]
\centering
\small
\setlength{\tabcolsep}{4pt}
\begin{tabular}{llrrrrrrrr}
\toprule
& & \multicolumn{4}{c}{\textbf{System-wise (\% OMR-NED)}} & \multicolumn{4}{c}{\textbf{Part-wise (\% OMR-NED)}}\\
\cmidrule(lr){3-6} \cmidrule(lr){7-10}
& & \multicolumn{2}{c}{Zeus} & \multicolumn{2}{c}{SMT} & \multicolumn{2}{c}{Zeus} & \multicolumn{2}{c}{SMT}\\
\cmidrule(lr){3-4} \cmidrule(lr){5-6} \cmidrule(lr){7-8} \cmidrule(lr){9-10}
\textbf{Format} & \textbf{Tokenization} & {Synth.} & {Scan.} & {Synth.} & {Scan.} & {Synth.} & {Scan.} & {Synth.} & {Scan.} \\
\midrule
LMXE & LMXE          & \phantom{0}\underline{\textbf{7.3}}\,(1.8) & \underline{\textbf{10.6}}\,(3.1) & \phantom{0}\underline{\textbf{5.2}}\,(1.0) & \underline{\textbf{11.3}}\,(2.5) & \phantom{0}\underline{\textbf{3.6}}\,(1.3) & \phantom{0}\underline{\textbf{5.9}}\,(2.3) & \phantom{0}\underline{\textbf{4.2}}\,(0.8) & \underline{\textbf{10.4}}\,(3.4) \\
     & LMXE-P        & \phantom{0}7.8\,(1.9) & 11.0\,(2.6) & \phantom{0}\underline{\textbf{5.2}}\,(1.2) & 13.0\,(3.8) & ---\phantom{0.0} & ---\phantom{0.0} & ---\phantom{0.0} & ---\phantom{0.0} \\
\midrule
\texttt{**kern} & EKERN  & 25.2\,(6.9) & 29.9\,(6.8) & 13.6\,(4.6) & 19.9\,(6.2) & \phantom{0}5.5\,(1.7) & \phantom{0}7.4\,(2.2) & \phantom{0}6.3\,(1.9) & \textbf{11.5}\,(3.0) \\
     & BEKERN        & \textbf{11.9}\,(2.8) & \textbf{16.5}\,(3.6) & \phantom{0}\textbf{7.8}\,(2.1) & \textbf{15.3}\,(4.4) & \phantom{0}\textbf{4.0}\,(1.5) & \phantom{0}\textbf{6.3}\,(2.2) & \phantom{0}\textbf{5.0}\,(1.6) & 11.7\,(2.6) \\
\midrule
ABC  & CABC          & \phantom{0}\textbf{8.6}\,(1.8) & \textbf{12.2}\,(2.6) & \phantom{0}\textbf{6.0}\,(1.2) & 18.5\,(6.9) & \phantom{0}\textbf{3.8}\,(1.2) & \phantom{0}\textbf{6.2}\,(1.9) & \phantom{0}\textbf{4.8}\,(1.4) & \textbf{11.4}\,(3.3) \\
     & BPE-256       & 15.0\,(2.5) & 20.1\,(3.9) & \phantom{0}7.6\,(1.3) & \textbf{13.7}\,(2.3) & \phantom{0}4.7\,(1.4) & \phantom{0}7.1\,(2.0) & \phantom{0}5.9\,(1.7) & 12.9\,(3.4) \\
     & BPE-512       & 23.0\,(2.9) & 28.7\,(4.6) & \phantom{0}9.3\,(1.2) & 15.2\,(3.7) & \phantom{0}6.2\,(1.4) & \phantom{0}9.4\,(2.4) & \phantom{0}7.0\,(1.7) & 13.0\,(2.4) \\
     & BPE-1024      & 34.7\,(2.6) & 40.6\,(3.8) & 10.6\,(1.3) & 16.5\,(3.8) & \phantom{0}7.9\,(1.6) & 11.7\,(2.2) & \phantom{0}8.0\,(1.4) & 13.9\,(4.4) \\
     & BPE-4096      & 62.4\,(1.7) & 64.7\,(1.8) & 13.3\,(2.6) & 19.6\,(4.3) & 18.1\,(1.7) & 23.3\,(3.1) & 10.7\,(2.1) & 17.1\,(3.4) \\
\bottomrule
\end{tabular}
\caption{OMR-NED (\%, mean (std) over four data splits) for all nine tokenization schemes evaluated. Lower is better. ABC-BPE rows are ordered by vocabulary size to show the trade-off between vocabulary granularity and accuracy. Bold values indicate the best results per metric, transcription method, data type, and encoding format, while underlined values denote the overall best results across metrics, methods, and data types.}
\label{tab:results}
\end{table*}

\subsection{Evaluation metric}\label{subsec:metric}
We evaluate all baselines with the OMR Normalized Edit Distance (OMR-NED)~\cite{omr-ned}, defined as
\begin{equation}
\mathrm{OMR\text{-}NED} = \frac{I + D}{N_1 + N_2},
\end{equation}
where $N_1$ and $N_2$ are the counts of musical symbols in the \texttt{music21}~\cite{music21} symbolic space for the prediction and the reference, respectively, and $I$ and $D$ are the insertion and deletion counts between them, computed in the same symbolic space via \texttt{musicdiff}~\cite{musicdiff}.\footnote{\url{https://github.com/gregchapman-dev/musicdiff}} OMR-NED is thus format-agnostic~\cite{transcoda}, enabling fair comparison across tokenization schemes with different vocabulary sizes and sequence lengths.

For system-level evaluation, OMR-NED is computed on the full predicted system sequence. For part-level evaluation, each of the four parts is predicted independently from its staff image, and OMR-NED is computed by pooling error counts across all four parts of the system:
\begin{equation}
\mathrm{OMR\text{-}NED}_{\mathrm{sys}} = \frac{\sum_{i=1}^{P}(I^{(i)} + D^{(i)})}{\sum_{i=1}^{P}(N_1^{(i)} + N_2^{(i)})},
\end{equation}
where $P$ is the number of parts in the system (4 for string quartets), so that part-level and system-level scores are directly comparable. For \texttt{**kern} and ABC, the original time signature of the system is injected at the first measure of each staff prediction to prevent parse failures during reconstruction; time-signature accuracy is therefore not comparable across segmentation levels for these two encodings.

%-----------------------------------------------------
\section{Baseline experiments}\label{sec:experiments}
%-----------------------------------------------------

\subsection{Training setup}

All models are trained with the AdamW~\cite{adamw} optimizer for 100{,}000 updates, with auto-regressive validation every 5{,}000 steps monitoring OMR-NED on the validation set. Batch sizes are 64 at the system level and 256 at the part level (reflecting the $\sim$4$\times$ larger sample count). Learning rates are set to 0.001 with a cosine schedule for Zeus and 0.0001 with no schedule for SMT.

Input images are converted to grayscale and padded with a value of 255: height is center-padded to a fixed 256 pixels (system) or 112 pixels (part), while width is right-padded dynamically to the widest sample of each mini-batch rather than to a fixed cap. Training uses the data augmentation scheme from~\cite{mayerZeus}.

Training is distributed across two NVIDIA RTX PRO 6000 Blackwell GPUs and four NVIDIA B200 GPUs. In total, 136 runs (34 configurations $\times$ 4 splits) are trained independently from scratch to convergence or up to 100{,}000 updates.

\subsection{Results and findings}
Table~\ref{tab:results} reports OMR-NED on synthetic and scanned test sets for all nine tokenization schemes, with the BPE rows ordered by vocabulary size to expose the trade-off between sequence length and accuracy.

\paragraph{Encoding choice substantially affects accuracy.} 
LMXE consistently achieves the lowest mean OMR-NED across both models and segmentation levels, followed closely by character-level ABC; \texttt{**kern}-based encodings sit a step behind, and byte-pair-encoded ABC variants are the weakest. Although SMT was originally proposed for \texttt{**kern}~\cite{rios2024sheet}, it performs better on LMXE than on either \texttt{**kern} tokenization here. All four ABC-BPE vocabularies underperform CABC, and Zeus degrades monotonically as BPE vocabulary grows; byte-pair encoding shortens sequences but does not, in this benchmark, improve transcription accuracy.

\paragraph{Part-level segmentation outperforms system-level.}
Part-level inputs yield lower OMR-NED than system-level inputs in every configuration. In principle, system-level inputs offer contextual cues unavailable at the part level, such as inferring time signatures from concurrent voices, deducing note durations from vertical alignment, or pitches from harmonic context. We did not observe evidence that the baselines exploit this context: error rates on the OMR-NED categories most directly tied to such cues (\texttt{note}, \texttt{notehead}) are no lower at the system level than at the part level. Time signatures show the same pattern under LMXE---the only encoding for which they are not injected (Section~\ref{subsec:metric})---with system-level error higher in 14 of the 16 model, test-set, and split combinations.

\paragraph{Robustness to domain shift.}
Synthetic-to-scanned transfer increases OMR-NED across all 34 configurations, but the gap varies sharply by model: Zeus degrades by an average of 39\% (3.6~pp) relative to its synthetic baseline, while SMT degrades by 100\% (6.7~pp).
Zeus is more robust than SMT to the synthetic-to-scanned shift; the architectural source of this gap is not isolated by these experiments.
The best scanned configuration is Zeus on LMXE at the part level (5.9\%).

\subsection{External baseline: Legato}\label{subsec:legato}
We additionally evaluate Legato~\cite{legato2026}, an ABC-based OMR model trained on PDMX~\cite{pdmx}, on the ABC-encoded test set of OSSQ-OMR using the released weights and tokenizer, without fine-tuning.\footnote{\url{https://huggingface.co/guangyangmusic/legato}} Two protocol choices differ from our other baselines: (i) we set the decoding cap to 2{,}048 / 1{,}024 tokens (system / part) to avoid premature termination under Legato's PDMX-trained tokenizer, and (ii) we use greedy rather than beam-search decoding for parity. The results of this study are reported in Table~\ref{tab:legato}. 

\begin{table}[!th]
\centering
\small
\begin{tabular}{lcc}
\toprule
\textbf{Source} & \textbf{System-wise} & \textbf{Part-wise} \\
\midrule
Synthetic & 36.1\,$\pm$\,3.2 & 51.2\,$\pm$\,2.5\\
Scanned & 66.3\,$\pm$\,2.7 & 74.5\,$\pm$\,3.1 \\
\bottomrule
\end{tabular}
\caption{OMR-NED (\%, mean\,$\pm$\,std across the four data splits) for Legato on the OSSQ-OMR test sets. Lower is better.}
\label{tab:legato}
\end{table}

As observed, Legato's system-wise OMR-NED on OSSQ-OMR (36.1\% synthetic, 66.3\% scanned) is close to its reported page-wise result on the 252-page OpenScore evaluation set in~\cite{legato2026} (32.9\%, 58.2\%); since OMR-NED is micro-averaged, the two are directly comparable, and the 3--8~pp gap is consistent with the change from beam to greedy decoding.
By contrast, our baselines trained on ABC encodings from OSSQ-OMR achieve substantially lower error than Legato (e.g., SMT-CABC at 6.0\% / 18.5\% system-wise vs.\ Legato's 36.1\% / 66.3\%; see Table~\ref{tab:results}), suggesting that ABC pretraining on PDMX alone is not sufficient for multi-part string-quartet OMR.

%-----------------------------------------------------
\section{Conclusions and future work}\label{sec:discussion}
%-----------------------------------------------------
We introduced OSSQ-OMR, the first dataset dedicated to multi-part OMR, together with a benchmark protocol and baseline results from two representative architectures. The benchmark shows that representation choices matter more than architecture: LMXE outperforms \texttt{**kern} and ABC under both models, part-level inputs outperform system-level ones in every configuration, and byte-pair compression of ABC degrades accuracy monotonically with vocabulary size. The best configuration reaches 3.6\% OMR-NED on synthetic and 5.9\% on scanned inputs, confirming that multi-part OMR is feasible while leaving clear headroom on scanned sources.

The results and insights obtained in this work open several promising directions for future research in OMR. In particular, advancing beyond the current scope will likely require the development of richer multi-part benchmarks, including settings with larger instrumental ensembles such as orchestral scores. Beyond data, we also see full-page end-to-end recognition as a key direction for scaling current approaches, potentially enabling more unified models that better capture global score structure.

%-----------------------------------------------------
\section{Acknowledgments}\label{sec:ack}
%-----------------------------------------------------

This work was supported by the Ministry of Education of the Republic of Korea and the National Research Foundation of Korea (NRF-2024S1A5C3A03046168), and by the ``Advanced GPU Utilization Support Program'' funded by the Government of the Republic of Korea (Ministry of Science and ICT).

%-----------------------------------------------------
\section{AI Usage Statement}\label{sec:ai}
%-----------------------------------------------------
AI tools were used only for minor language editing, such as paraphrasing for readability and grammar checking. All dataset construction, experimental design, training, evaluation, and analysis were carried out by the authors, who verified all reported results.

% For BibTeX users:
\bibliography{ISMIRtemplate}

\end{document}